\documentclass[runningheads]{llncs}
\usepackage[T1]{fontenc}
\usepackage{orcidlink}
\usepackage{adjustbox}
\usepackage{marvosym}
\usepackage{enumitem}

\usepackage{mathtools}
\usepackage[linesnumbered,ruled,vlined]{algorithm2e}
\SetKwInput{KwInput}{\textbf{Input}}
\SetKwInput{KwOutput}{\textbf{Output}}

\usepackage{hyperref}
\hypersetup{
    colorlinks=true, 
    linkcolor=blue, 
    citecolor=blue,
    urlcolor=blue, 
}
\usepackage{graphicx}
\begin{document}
\title{Efficient Hyperparameter Importance Assessment for CNNs}
%
%
\author{Ruinan Wang\inst{1}\orcidlink{0009-0003-1218-2921} \and
Ian Nabney\inst{2}\textsuperscript{(\Letter)} \orcidlink{0000-0003-1513-993X}\and
Mohammad Golbabaee\inst{3}\orcidlink{0000-0001-5822-2990}}
\authorrunning{R. Wang et al.}
%
\institute{University of Bristol, Bristol, United Kingdom \\
\email{\{zg21696,in17746,an22148\}@bristol.ac.uk}}
\maketitle              
\begin{abstract}
Hyperparameter selection is an essential aspect of the machine learning pipeline, profoundly impacting models' robustness, stability, and generalization capabilities. Given the complex hyperparameter spaces associated with Neural Networks and the constraints of computational resources and time, optimizing all hyperparameters becomes impractical. In this context, leveraging hyperparameter importance assessment (HIA) can provide valuable guidance by narrowing down the search space. This enables machine learning practitioners to focus their optimization efforts on the hyperparameters with the most significant impact on model performance while conserving time and resources. This paper aims to quantify the importance weights of some hyperparameters in Convolutional Neural Networks (CNNs) with an algorithm called N-RReliefF, laying the groundwork for applying  HIA methodologies in the Deep Learning field. We conduct an extensive study by training over ten thousand CNN models across ten popular image classification datasets, thereby acquiring a comprehensive dataset containing hyperparameter configuration instances and their corresponding performance metrics. It is demonstrated that among the investigated hyperparameters, the top five important hyperparameters of the CNN model are the number of convolutional layers, learning rate, dropout rate, optimizer and epoch.

\end{abstract}

\section{Introduction}
With the growing prominence of Deep Learning and Automated Machine Learning frameworks, Hyperparameter Optimization (HPO) techniques have evolved from manual, empirical tuning to automated methods such as Random Search~\cite{bergstra2012random}, Bayesian Optimization~\cite{snoek2012practical}, and Evolutionary Algorithms~\cite{nannen2007efficient}. However, optimizing all hyperparameters in large search spaces is often impractical due to limited computational resources and time. Furthermore, regardless of the HPO algorithm, we must manually define the hyperparameter search space~\cite{frazier2018tutorial}, often relying on rules of thumb that may lack rigour. Hyperparameter Importance Assessment (HIA)~\cite{hutter2014efficient} can guide users by focusing on the most impactful hyperparameters. However, its use in Deep Learning remains underexplored due to the variety of hyperparameters and the challenge of collecting performance data across numerous configurations.

To address this gap, this study aims to investigate whether an HIA method called N-RReliefF~\cite{sun2019hyperparameter} can offer insights into the importance of some hyperparameters for Convolutional Neural Networks (CNNs). By introducing HIA into the realm of Deep Learning, our study could enhance the understanding of the intricate workings within Neural Network models, often perceived as "black boxes". Although our approach doesn't directly explain the link between input data and model outputs, it enhances model transparency by revealing the influence of hyperparameters on performance, thereby assisting model developers in making more informed decisions during the model development process. The main contributions of this paper are as follows:

\begin{enumerate}[label={\arabic{enumi})}]
    \item Train over 10000 CNN models on 10 image classification datasets and record their hyperparameter configurations and their corresponding performance data, which also could be used for future studies, e.g., analysis of Model Complexity and Efficiency and architectural analysis.
    \item Use N-RRelifF to assess the individual importance of 11 hyperparameters in CNN, generating a ranking based on the importance weights.
    \item Evaluate the importance weights of pairs of investigated hyperparameters.
    \item Explore further the importance weights of the hyperparameters having the dependent relationship ("the number of filters in different convolutional layers" and "the number of layers").
\end{enumerate}

The remainder of this paper is organized as follows: Section \ref{sec: related_work} reviews some related works, focusing on the development of HIA and its application to various machine learning algorithms. Section \ref{sec: derivation} provides a detailed formula derivation of the N-RReliefF algorithm. Section \ref{sec: experiments} outlines the specific details of the HIA experiment. Section \ref{sec: evaluation} presents the comprehensive evaluation and a series of analyses for HIA results while Section \ref{sec: conclusions} concludes the paper and outlines future research directions.

\section{Related Works}
\label{sec: related_work}
When mentioning how to quantify the importance of hyperparameters, another similar field that has many practical approaches probably comes to mind, i.e., Feature Selection~\cite{kumar2014feature}. Feature Selection is the process of reducing the dimensionality of input features when developing predictive models to save the computational cost of modelling and, in some cases, improve model performance~\cite{zebari2020comprehensive}. It can use statistical measures to score the correlations between each input feature and the model performance for selecting the most relevant features~\cite{zebari2020comprehensive}, which is very similar to the aim of HIA. It was found that prior research has applied feature selection methodologies to the HIA of traditional machine learning models.

At the start of HIA, Bartz-Beielstein et al.\cite{bartz2006new} used contour visualization to explore interactive parameters. However, this method cannot handle the configuration space formed by discrete hyperparameters in the algorithm configuration scene. Discrete hyperparameters refer to parameters that take on discrete values, such as choosing different optimizers, activation functions, or network architectures. The discrete nature of these parameters makes it difficult to apply traditional methods that assume a continuous space. In 2007, Nannen et al.\cite{nannen2007efficient} proposed an evolutionary algorithm for parameter correlation estimation. This method assumes a smooth hyperparameter response surface and can handle continuous hyperparameters but is still limited in dealing with many hyperparameter configurations. In order to solve the problem of discrete hyperparameters, Hutter et al.~\cite{hutter2014efficient} used model-based techniques to study the importance of hyperparameters and hyperparameter interactions. They proposed forward selection algorithms and Functional Analysis of Variance (ANOVA) algorithms. The forward selection algorithm iteratively adds greedy hyperparameters with minimum root mean square error in the validation set to build regression models iteratively~\cite{worland2018improving}. Functional ANOVA (FANOVA) applies variance decomposition to random forest models to assess hyperparameter importance~\cite{hutter2014efficient}. These algorithms are excellent in dealing with the high dimensionality and dispersion of the algorithm configuration space, but they require iterative model construction, which leads to increased time complexity. Besides, this study only evaluates the importance of hyperparameters on one single specific dataset. Therefore, in another paper, Rijn et al.~\cite{van2017empirical} made an empirical study to obtain more representative results. They applied this method to 100 datasets to determine the most important hyperparameters of random forest and AdaBoost.

To mitigate the rise in time complexity introduced by modelling and to understand the order of the hyperparameter importance of the algorithm itself, Sun et al.~\cite{sun2019hyperparameter} proposed an algorithm called N-RReliefF, which is an extension of the Relief family algorithms. Sun et al.~\cite{sun2019hyperparameter} applied N-RReliefF, Forward Selection and Functional ANOVA to evaluate the importance of some hyperparameters from SVM and random forest classifiers. The final results indicate that the hyperparameter importance rankings produced by these three methods are consistent. For SVM, ``gamma" is the most important hyperparameter, ``complexity" is the second most important hyperparameter, and ``imputation" is the least important. For Random Forest, ``split criterion" and ``bootstrap" are the first two most important hyperparameters, and ``imputation" is the least important. Additionally, the experiments revealed that N-RReliefF requires significantly less computational time than the other two methods, highlighting its efficiency advantage without compromising the quality of the results.

\section{Algorithm Derivation}
\label{sec: derivation}

\subsection{Notation}
\begin{itemize}[label={$\bullet$}]
    \item For a machine learning model $f$, $\left[\Theta\right] \coloneqq \left\{ \Theta_1, \ \Theta_2,\dots,\ \Theta_k \right\}$ represents the hyperparameter configuration space where $\Theta_1$ stands for the first hyperparameter, and so forth. In this space, an instance $h_m$ $(h_m\in\Theta)$ can be denoted as a vector $h_m \coloneqq \left(\theta_{m_1}, \ \theta_{m_2}, \ \dots, \ \theta_{m_k}\right)$ where $\theta_{m_1}$ stands for the value of the first hyperparameter in the instance $h_m$.
    
    \item $[H] \coloneqq \{h_1, \ h_2, \dots, \ h_n\}$ is defined as the collection of hyperparameter configuration instances.    
    
    \item Utilizing each instance from $H$, with the same training dataset $D_{train}$ and the same test dataset $D_{test}$, a corresponding collection of performance metric $\left[P\right]$ can be obtained: 
      \\ $\left[P\right] \coloneqq \left\{f(h_1, D_{train}, D_{test}), f(h_2, D_{train}, D_{test}),\dots, f(h_n, D_{train}, D_{test})\right\} =\left\{p_1, p_2, \dots, p_n\right\}$
    \item The input dataset $D$ for N-RReliefF can be represented as\\ $\left[D\right] \coloneqq \left\{\left(h_1, \ p_1\right),\ \left(h_2, \ p_2\right), \dots, \ \left(h_n, \ p_n\right)\right\}$.
    
    \item The distance between the two hyperparameter configuration instances is calculated using the Euclidean distance: $dist(h_m, h_j) \coloneqq \sqrt{\sum_{i=1}^{k} (\theta_{mi}-\theta_{ji})^2}$.
    
    \item $h_{\text{NN}_j}$ represents the $j$-th nearest neighbor to the randomly sampled instance $h_{m}$. 
    \item Correspondingly, $p_{\text{NN}_j}$ and $p_{m}$ represent the performance metrics associated with $h_{\text{NN}_j}$ and $h_{m}$.
    \item $\text{rank}(h_m, \ h_{\text{NN}_j})$ calculates the rank of $h_{\text{NN}_j}$ among the first $J$ nearest neighboring instances to $h_m$.
    
    \item $\text{diff}(\theta_{m_1}, \ \theta_{\text{NN}_{j_1}})$ denotes the difference between the values of the hyperparameter $\Theta_1$ for $h_m$ and $h_{\text{NN}_j}$. If $\Theta_1$ is numerical: $$\text{diff}(\theta_{m_1}, \ \theta_{\text{NN}_{j_1}}) \coloneqq \left| \frac{\theta_{m_1}-\theta_{\text{NN}_{j_1}}}{\max(\theta_1)-\min(\theta_1)} \right|$$

    $\max(\theta_1)$ means the maximum value of $\Theta_1$ on all collected hyperparameter configuration instances.
      
    And if $\Theta_1$ is non-numerical: $$\text{diff}(\theta_{m_1}, \ \theta_{\text{NN}_{j_1}}) \coloneqq \begin{cases} 0, & \text{if } \theta_{m_1} = \theta_{\text{NN}_{j_1}} \\ 1, & \text{otherwise} \end{cases}$$
    
    \item $\text{diff}(p_m, p_{\text{NN}_j})$ denotes the difference between two model performances $p_m$ and $p_{\text{NN}_j}$ under the corresponding instances $h_m$ and $h_{\text{NN}_j}$: $\text{diff}(p_m, \ p_{\text{NN}_j}) \coloneqq \rvert p_m - p_{\text{NN}_j} \rvert$.

    \item In the current context, whether calculating the differences between hyperparameters or the differences between model performances, we should account for neighbouring instances closer to $h_m$ should have a higher degree of influence on the results. Therefore, a weight term $d(h_m, h_{\text{NN}_j})$ is introduced:
    $$d(h_m, h_{\text{NN}_j}) = \frac{d'(h_m, h_{\text{NN}_j})}{\sum_{j=1}^{J} d'(h_m, h_{\text{NN}_j})}$$
    $$d'(h_m, h_{\text{NN}_j}) = e^{-\left(\frac{\text{rank}(h_m, h_{\text{NN}_j})}{\sigma}\right)^2}$$
    The influence of the neighbouring instance exponentially decreases as its distance rank among the first $J$ neighbouring instances increases. $\sigma$ is a user-defined parameter for controlling the extent of the influence of the rank on the result~\cite{robnik1997adaptation}. The rationale behind employing ranks rather than actual distances is that utilizing ranks standardizes the influence each instance has on the weight calculations, ensuring that the nearest instances—and those that follow—consistently exert the same level of impact regardless of the dataset's peculiarities~\cite{robnik1997adaptation}. $d(h_m, h_{\text{NN}_j})$ is actually a normalization of $d'(h_m, h_{\text{NN}_j})$, allowing us to interpret $d'(h_m, h_{\text{NN}_j})$ probabilistically.
 
    \item $W\left[ \Theta \right]$ represents a vector of the importance weight of the investigated hyperparameters. This is the output for N-RReliefF and the calculation process and details will be provided later.
\end{itemize}

\subsection{Estimation of \texorpdfstring{$W\left[ \Theta \right]$}{Importance Weights}  in the Probabilistic Framework}
The key idea of the Relief family of algorithms is to estimate the quality of an attribute (i.e., the influence of hyperparameters on the model performance metric) by assessing how well the attribute values (i.e., hyperparameters) distinguish the outputs of the nearest neighbour instances~\cite{robnik2003theoretical}. Relief's estimate of $W[\Theta_k]$ can be written as the approximation of the difference between these two probabilities~\cite{kononenko1994estimating}:
\begin{align}
W[\Theta_k] := \ &P(\text{diff.} \Theta_k \mid \text{nearest diff.class}) - P(\text{diff.} \Theta_k \mid \text{nearest same class})\label{eq:formula1}
\end{align}

In Eq.\ref{eq:formula1}, the first term quantifies the degree of difference in the hyperparameter $\Theta_k$ values when comparing an instance with its nearest neighbour from a different class. Conversely, the second term measures the degree of difference in $\Theta_k$ values for that instance and its nearest neighbour from the same class.

However, Relief was designed under the assumption that the model outputs are discrete categories. In reality, the performance metric $P$ is a continuous variable so the notion of "the same class" and "the different class" does not apply in HIA. To address this challenge in regression problems, a variant known as RReliefF was proposed~\cite{robnik1997adaptation}. Unlike its predecessor, RReliefF doesn't rely on exact knowledge of whether two instances belong to the same class. Instead, it adopts a probabilistic approach to quantify the differences in model outputs, leading to a need for reformulating $W[\Theta_k]$ for this context. The following will derive the revised formulation of $W[\Theta_k]$ in the RReliefF framework. Based on Eq.\ref{eq:formula1}, we can rewrite $W[\Theta_k]$ to form Eq.\ref{eq:formula2} for regression problems.
\begin{align}
W[\Theta_k] := \ &P(\text{diff. } \Theta_k \mid \text{nearest diff. } p) - P(\text{diff. } \Theta_k \mid \text{nearest same } p)\label{eq:formula2}
\end{align}

Given that the model outputs are continuous variables, we can assess the variability in the output $P$ for a given instance relative to its nearest neighbours within a specified range. This variability is quantified by Eq.\ref{eq:formula3}, which represents observing the difference degree in the model output values among neighbouring instances.
\begin{align}
P_{\text{diff}(p)} := P(\text{diff. } p \mid \text{nearest instances}) \label{eq:formula3}
\end{align}

Meanwhile, we also can get Eq.\ref{eq:formula4}, the probability of the difference degree in the values of the hyperparameter $\Theta_k$ when comparing one instance with all its nearest neighbours within a specific range. This probability quantifies the degree of variation in $\Theta_k$ across neighbouring instances.
\begin{align}
P_{\text{diff}(\Theta_k)} := P(\text{diff. } \Theta_k \mid \text{nearest instances})  \label{eq:formula4}
\end{align}

Furthermore, we can define another important conditional probability with Eq.\ref{eq:formula5}. This quantifies the probability of a change in \( P \), conditional upon differences in the hyperparameter \( \Theta_k \) within the nearest instances. It specifically shows how variability in the hyperparameter is associated with variability in the model output.
\begin{align}
P_{\text{diff}(p)\mid \text{diff}(\Theta_k)} := P(\text{diff. } p \mid \text{diff. } \Theta_k,  \text{ nearest instances})
\label{eq:formula5}
\end{align}

Based on Bayes' Theorem, we can get the first term of Eq.\ref{eq:formula2}:
\begin{align}
P(&\text{diff. } \Theta_k \mid \text{nearest diff. } p) = \frac{ P_{\text{diff}(p)\mid \text{diff}(\Theta_k)} \cdot P_{\text{diff}(\Theta_k)}}{P_{\text{diff}(p)}}
\label{eq:formula6}
\end{align}

Within the probabilistic framework, we can acknowledge:
\begin{align}
P(&\text{same } p \mid \text{nearest instances}) + P_{\text{diff}(p)} = 1 \nonumber \\
P(&\text{same } p \mid \text{diff. } \Theta_k, \text{nearest}) + P_{\text{diff}(p)\mid \text{diff}(\Theta_k)}  = 1
\label{eq:formula7}
\end{align}

Eq.\ref{eq:formula7} can lead us to derive the second term of Eq.\ref{eq:formula2}:
\begin{align}
P(&\text{diff. } \Theta_k \mid \text{nearest same } p) = \frac{P(\text{same } p \mid \text{diff. } \Theta_k, \text{nearest}) \cdot P_{\text{diff}(\Theta_k)}}{P(\text{same } p \mid \text{nearest instances})} \nonumber \\
&= \frac{ (1 - P_{\text{diff}(p)\mid \text{diff}(\Theta_k)}) \cdot P_{\text{diff}(\Theta_k)}}{1 - P_{\text{diff}(p)}}
\label{eq:formula8}
\end{align}

Considering that \( \text{diff}(p) \) and \( \text{diff}(\Theta_k) \) are not independent events:
\begin{align}
P_{\text{diff}(p) \text{ and } \text{diff}(\Theta_k)} = P_{\text{diff}(p)\mid \text{diff}(\Theta_k)}\cdot P_{\text{diff}(\Theta_k)}
\label{eq:formula9}
\end{align}

By combining these derived probabilities, the final representation of \( W[\Theta_k] \) in the RReliefF framework \( W[\Theta_k] \) would be:
\begin{align}
W[\Theta_k] =  &\frac{ P_{\text{diff}(p)\mid \text{diff}(\Theta_k)}\cdot P_{\text{diff}(\Theta_k)}}{P_{\text{diff}(p)}} - \frac{ (1 - P_{\text{diff}(p)\mid \text{diff}(\Theta_k)})\cdot P_{\text{diff}(\Theta_k)}}{1 - P_{\text{diff}(p)}} \nonumber \\
= &\frac{ P_{\text{diff}(p) \text{ and } \text{diff}(\Theta_k)}}{P_{\text{diff}(p)}} - \frac{ P_{\text{diff}(\Theta_k)} - P_{\text{diff}(p) \text{ and } \text{diff}(\Theta_k)}}{1 - P_{\text{diff}(p)}}
\label{eq:formula10}
\end{align}

After performing the above process on all hyperparameters, the importance weights of all hyperparameters, $W[\Theta]$, can be obtained. In addition, to compute the importance weights for combinations of hyperparameters, we applied an enhanced normalization formula. This approach is designed to scale the weights in a manner that considers the exponential of the sum of individual hyperparameter weights, thus facilitating comparative analysis of their combined influence~\cite{sun2019hyperparameter}. The improved normalization formula is expressed as follows:
\begin{align}
W[\Theta_{m} \& \Theta_{n}] = \frac{e^{W(\Theta_{m}) + W(\Theta_{n})}}{e^{\sum W(\Theta)}}
\label{eq:formula11}
\end{align}

\subsection{Approximating Key Terms in N-RReliefF}

After completing the derivation of the N-RReliefF formula, it is found that to estimate \( W[\Theta] \) in Eq.\ref{eq:formula10}, we only need to approximate three terms: Eq.\ref{eq:formula3}, Eq.\ref{eq:formula4}, and Eq.\ref{eq:formula9}. Three weights, \( N_{\text{diff}(p)} \), \( N_{\text{diff}(\Theta_k)} \), and \( N_{\text{diff}(p) \text{ and } \text{diff}(\Theta_k)} \) are defined as the approximation values of these three terms. 

$N_{\text{diff}(p)}$ indicates the cumulative difference situation between the randomly sampled instance's performance metric, $p_m$, and each neighbouring instance's performance metric $p_{\text{NN}_j}$ where the number of neighbouring instances is $J$.
\begin{align}
N_{\text{diff}(p)} = \sum_{j=1}^{J} \text{diff}(p_m, p_{\text{NN}_j}) \cdot d(h_m, h_{\text{NN}_j})
\label{eq:formula12}
\end{align}

$N_{\text{diff}(\Theta_k)}$ indicates the accumulation of differences on the specific hyperparameter $\Theta_k$ between the randomly sampled instance, $h_m$, and its every neighbouring instance, $h_{\text{NN}_j}$.
\begin{align}
N_{\text{diff}(\Theta_k)} = \sum_{j=1}^{J} \text{diff}(\theta_{m_k}, \theta_{\text{NN}_{j_k}}) \cdot d(h_m, h_{\text{NN}_j})
\label{eq:formula13}
\end{align}
\noindent $N_{\text{diff}(p) \text{ and } \text{diff}(\Theta_k)}$ simultaneously accounts for the cumulative differences in both the performance metric $p$ and a specific hyperparameter $\Theta_k$ between a randomly sampled instance $h_m$ and each of its neighbouring instances $h_{\text{NN}_j}$.
\begin{align}
N_{\text{diff}(p) \text{ and } \text{diff}(\Theta_k)} = \sum_{j=1}^{J} \text{diff}(p_m, p_{\text{NN}_j}) \cdot \text{diff}(\theta_{m_k}, \theta_{\text{NN}_{j_k}}) \cdot d(h_m, h_{\text{NN}_j})\label{eq:formula14}
\end{align}

The implementation process of N-RReliefF is outlined in Algorithm \ref{alg:N-RReliefF}.

\begin{adjustbox}{width=\textwidth, center}
\begin{algorithm}[H]
\SetAlgoLined
\KwInput{The collection of hyperparameter configuration instances and corresponding performance metrics: $[D] \coloneqq \left\{\left(h_1, \ p_1\right),\ \left(h_2, \ p_2\right),\ \dots, \ \left(h_n, \ p_n\right)\right\}$; The iteration times: $m$ (user-defined); The number of neighbouring instances: $J$ (user-defined).}
\KwOutput{The importance weight vectors for individual hyperparameters, $W[\Theta]$, and for hyperparameter combinations, $W(\Theta_{m} \&\Theta_{n})$.}
\BlankLine

Initialization: set weights $N_{\text{diff}(p)}$, $N_{\text{diff}(\Theta_k)}$, $N_{\text{diff}(p) \text{ and } \text{diff}(\Theta_k)} = 0$ \;
\For{m = 1 \KwTo M}{
    randomly sample $(h_m, \ p_m)$ from $[D]$\;
    find the $J$ neighbouring instances, $[R] = \left\{(h_{NN_1},\ p_{NN_1}), (h_{NN_2},\ p_{NN_2}),\dots,(h_{NN_J},\ p_{NN_J}) \right\}$

    \For{ j = 1 \KwTo J}{
    $N_{\text{diff}(p)}= N_{\text{diff}(p)} + \text{diff}(p_m, p_{NN_j}) \cdot d(h_m, h_{NN_j})$\;
    \For{ k = 1 \KwTo the number of hyperparameters $K$}{
    $N_{\text{diff}(\Theta_k)} = N_{\text{diff}(\Theta_k)} + \text{diff}(\theta_{m_k}, \theta_{{NN_j}_k}) \cdot d(h_m, h_{NN_j})$\;
    $N_{\text{diff}(p) \text{ and } \text{diff}(\Theta_k)} = N_{\text{diff}(p) \text{ and } \text{diff}(\Theta_k)} + \text{diff}(p_m, p_{NN_j}) \cdot \text{diff}(\theta_{m_k}, \theta_{{NN_j}_k}) \cdot d(h_m, h_{NN_j})$\;
        }
    }
}
$N_{\text{diff}(p)} = \frac{N_{\text{diff}(p)}}{M}$\;
\For{ k = 1 \KwTo $K$}{
$N_{\text{diff}(\Theta_k)} = \frac{N_{\text{diff}(\Theta_k)}}{M}$\;
$N_{\text{diff}(p) \text{ and } \text{diff}(\Theta_k)} = \frac{N_{\text{diff}(p) \text{ and } \text{diff}(\Theta_k)}}{M}$\;
$W[\Theta_k] = \frac{ N_{\text{diff}(p) \text{ and } \text{diff}(\Theta_k)}}{N_{\text{diff}(p)}} - \frac{ N_{\text{diff}(\Theta_k)} - N_{\text{diff}(p) \text{ and } \text{diff}(\Theta_k)}}{1 - N_{\text{diff}(p)}}$\;
}

\For{ m = 1 \KwTo $K$}{
\For{ n = 1 \KwTo $K$}{
$W[\Theta_{m} \& \Theta_{n}] = \frac{e^{W(\Theta_{m}) + W(\Theta_{n})}}{e^{\sum W(\Theta)}}$
}
}
\caption{N-RReliefF Algorithm in HIA}
\label{alg:N-RReliefF}
\end{algorithm}
\end{adjustbox}

\section{Experimental Setup}
\label{sec: experiments}

\subsection{Implementation Procedures}

\begin{figure}[!ht]
    \caption{\textbf{Workflow Diagram in Hyperparameter Importance Assessment}}
    \centering
    \includegraphics[scale = 0.3]{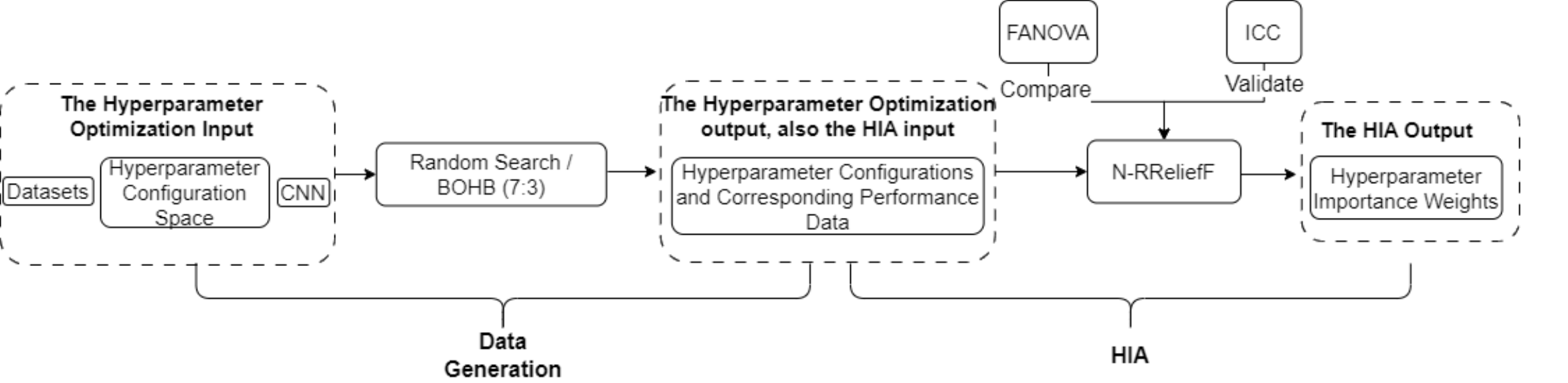}
    \label{fig: HIA Flow}
\end{figure}

The experimental procedure is shown in Figure \ref{fig: HIA Flow}, along with detailed explanations of how each step is performed. All experiments were conducted on a machine equipped with an NVIDIA GeForce RTX 3070Ti GPU, a 12th Gen Intel(R) Core(TM) i7-12700KF processor (3.60 GHz), and 32 GB of RAM.

The first step is data generation, using accuracy as the primary metric. We trained over 10,000 CNN models across ten image classification datasets, employing both Random Search and BOHB (Bayesian Optimization and Hyperband)~\cite{falkner2018bohb} in a 7:3 ratio to ensure a balanced performance data distribution. Random Search tends to focus on lower-performance areas, while BOHB quickly identifies high-performance regions. This hybrid approach enabled efficient data generation, supporting a robust hyperparameter importance assessment.

Subsequently, exploratory experiments were conducted to evaluate the importance weights of the individual and joint hyperparameters using N-RReliefF (Algorithm \ref{alg:N-RReliefF}). We also fixed the number of convolutional layers to assess the effect of filter counts.

For result verification, we applied a repeated experiment strategy, generating 10 subsets through random sampling. These subsets were analyzed using Intraclass Correlation Coefficients (ICC) ~\cite{von2009reliability}, with values between 0.75 and 0.90 confirming the robustness of our results~\cite{kim2017value}. Additionally, FANOVA~\cite{hutter2014efficient} is utilized for comparative analysis with N-RReliefF, further validating the reliability of results.

\subsection{Hyperparameter Configuration Space and Network Structure of CNNs}
The experiment examined the individual and joint importance of 11 hyperparameters, noting that some have dependent relationships, such as the number of convolutional layers and the kernels per layer. Dependent hyperparameters can't be analyzed individually alongside those that influence them. Thus, we propose fixing "parent" hyperparameters when studying "child" hyperparameters. For example, to compare kernel counts across layers, we first fix the number of layers. Table \ref{fig:hyperparameter} lists the hyperparameters, their data types, configuration spaces, and default values.
\begin{table}
 \caption{\textbf{Configuration Space of Investigated Hyperparameters in CNN}}
\centering
\begin{adjustbox}{scale = 0.9}
\begin{tabular}{|p{5.5cm}|c|c|c|}   
\hline {\textbf{Hyperparameter}} & 
			{\textbf{Data Type}} & 
			{\textbf{Configuration Space}}&
			{\textbf{Default Value}}\\ 
			\hline   batch size & integer & {32, 64, 128, 256} & 32\\ 
			\hline   dropout rate & float & [0.0, 0.9] & 0.5\\ 
			\hline   epoch & integer & [1, 10] & 2\\  
			\hline   the number of input channels & integer & {1, 3} & 0\\  
			\hline   convolutional kernel size & integer & [2, 3] & 2\\ 
			\hline   learning rate & float & [1e-6, 1e-1] & 1e-2\\ 
			\hline num\_conv\_layers & integer & [1, 3] & 2\\
			\hline   the number of filters in fully connected & integer & [8, 256] & 32\\
			\hline   optimizer & categorical & {Adam, SGD} & Adam\\ 
			\hline  padding & integer & {0, 1} & 0\\ 
			\hline  stride & integer & [1, 2] & 1\\  
			\hline the number of filters in cov layer 1 & integer & [4, 64] & 16\\
			\hline the number of filters in cov layer 2 & integer & [4, 64] & 16\\ 
			\hline the number of filters in cov layer 3 & integer & [4, 64] & 16\\ 
			\hline
\end{tabular} 
\end{adjustbox}
\label{fig:hyperparameter}
\end{table}

\subsection{Network Structure}
The structure of the CNN model changes dynamically during the data generation phase, but several details are fixed: (1) Each convolutional layer is followed by a ReLU activation function and a pooling layer. When the structural hyperparameter, the number of convolutional layers, is greater than one, additional ReLU and pooling layers are added after each convolutional layer. (2) Each pooling layer consistently uses a max-pooling strategy. (3) The final output is generated using a Softmax function. (4) Before the Softmax function, there are two identical combinations arranged sequentially, each consisting of a dropout layer followed by a fully connected layer, where the hyperparameter, dropout rate, is shared between both dropout layers. (5) The first fully connected layer is followed by a ReLU function. (6) Unless otherwise specified, hyperparameters such as padding, stride, or dropout rate apply consistently across all relevant layers.

\subsection{Selected Datasets}
\begin{table}[!ht]
	\centering\normalsize
 	\caption{\textbf{the Selected Image Datasets}}
  \begin{adjustbox}{scale = 0.6}
\begin{tabular}{|c|c|c|c|c|}   
\hline {\textbf{Dataset Name}} & 
			{\textbf{Data Amount}} & 
			{\textbf{Class Num}}&
			{\textbf{Channel Num}} &
			{\textbf{Size}}\\ 
			\hline   CIFAR-10 & 60000 & 10 & 3 & 32*32\\ 
			\hline   CIFAR-100 & 60000 & 100 & 3 & 32*32\\ 
			\hline   MNIST & 70000 & 10 & 1 & 28*28\\  
			\hline   EMNIST & 131600 & 47 & 1 & 28*28 \\  
			\hline   Fashion-MNIST & 70000 & 10 & 1 & 28*28\\ 
			\hline   EuroSAT & 37000 & 10 & 3 & 64*64\\ 
			\hline   SEMEION & 1593 & 10 & 3 & 16*16\\  
			\hline   STL10 & 13000 & 10 & 3 & 96*96\\ 
			\hline   SVHN & 99289 & 10 & 3 & 32*32 \\  
			\hline   USPS & 9298 & 10 & 1 & 16*16 \\ 
			\hline
\end{tabular}   
\end{adjustbox}
     \label{fig:selectedDatasets}
\end{table}
Given that the model to be evaluated is a CNN, which is most commonly used for image classification, we selected ten classic and widely used benchmark image classification datasets from publicly available sources to generate the hyperparameter configurations and the corresponding performance data. These datasets are chosen to represent a variety of scenarios. To assess the impact of the number of input channels on the CNN model's performance, the datasets include five colour and five grayscale collections. Table \ref{fig:selectedDatasets} provides specific information about each dataset. During the HIA input data generation phase, we adhered to the conventional practice of splitting the training and validation sets of each dataset in an 8:2 ratio.

\section{Evaluation and Results}
\label{sec: evaluation}

\subsection{Initial Data Exploration}
\begin{figure}[!ht]
    \centering
    \begin{minipage}[b]{0.48\linewidth} 
        \centering
        \caption{\textbf{The Amount of HIA Inputs Generated from Datasets}}
        \begin{adjustbox}{width=\textwidth, center}
        \begin{tabular}{|c|c|}   
            \hline 
            {\textbf{Dataset}} & {\textbf{The Amount of Data}}\\ 
            \hline   CIFAR10 & 1000 \\ 
            \hline   CIFAR100 & 1000\\ 
            \hline   EMNIST & 1150\\  
            \hline   EuroSAT & 1000 \\  
            \hline   FashionMNIST & 1000 \\ 
            \hline   MNIST & 1500\\ 
            \hline   SEMEION & 1000\\  
            \hline   STL10 & 1000 \\  
            \hline   SVHN & 1000\\ 
            \hline   USPS & 1500\\  
            \hline
        \end{tabular}
        \end{adjustbox}
        \label{tab:dataamount}
    \end{minipage}
    \hfill
    \begin{minipage}[b]{0.48\linewidth}
        \centering
        \caption{\textbf{Generated Data Distribution}}
        \includegraphics[width=\textwidth]{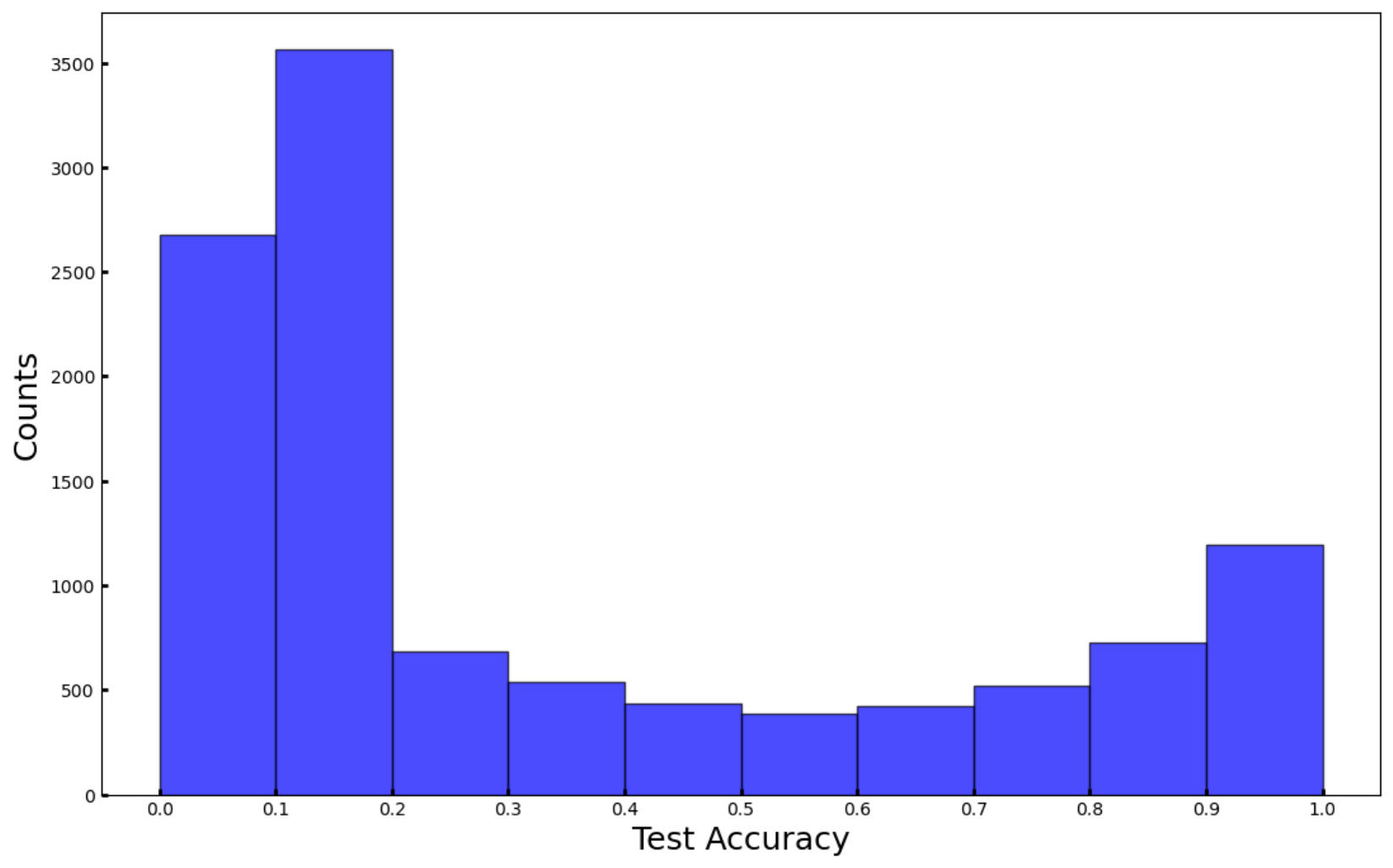}
        \label{fig:distribution}
    \end{minipage}
\end{figure}

In the initial phase of data exploration, we commenced with an examination of the volume of data generated from various image classification datasets. Figure \ref{tab:dataamount} illustrates that for each dataset, the quantity of hyperparameter configuration and associated performance data successfully surpassed the threshold of 1,000 instances. 

With further exploration into the data generated across all ten image classification datasets, the overall distribution of the data is illustrated. As depicted in Figure \ref{fig:distribution},  the distribution exhibits a bimodal tendency, skewing toward the extremes of performance, while the data volume within intermediate performance brackets remains comparatively sparse.

\subsection{Verifying the Reliability via ICC}

\begin{figure}[!ht]
    \caption{\textbf {Dot Plot about the Relative Positions of All Importance Weights}}
    \centering
    \includegraphics[scale=0.32]{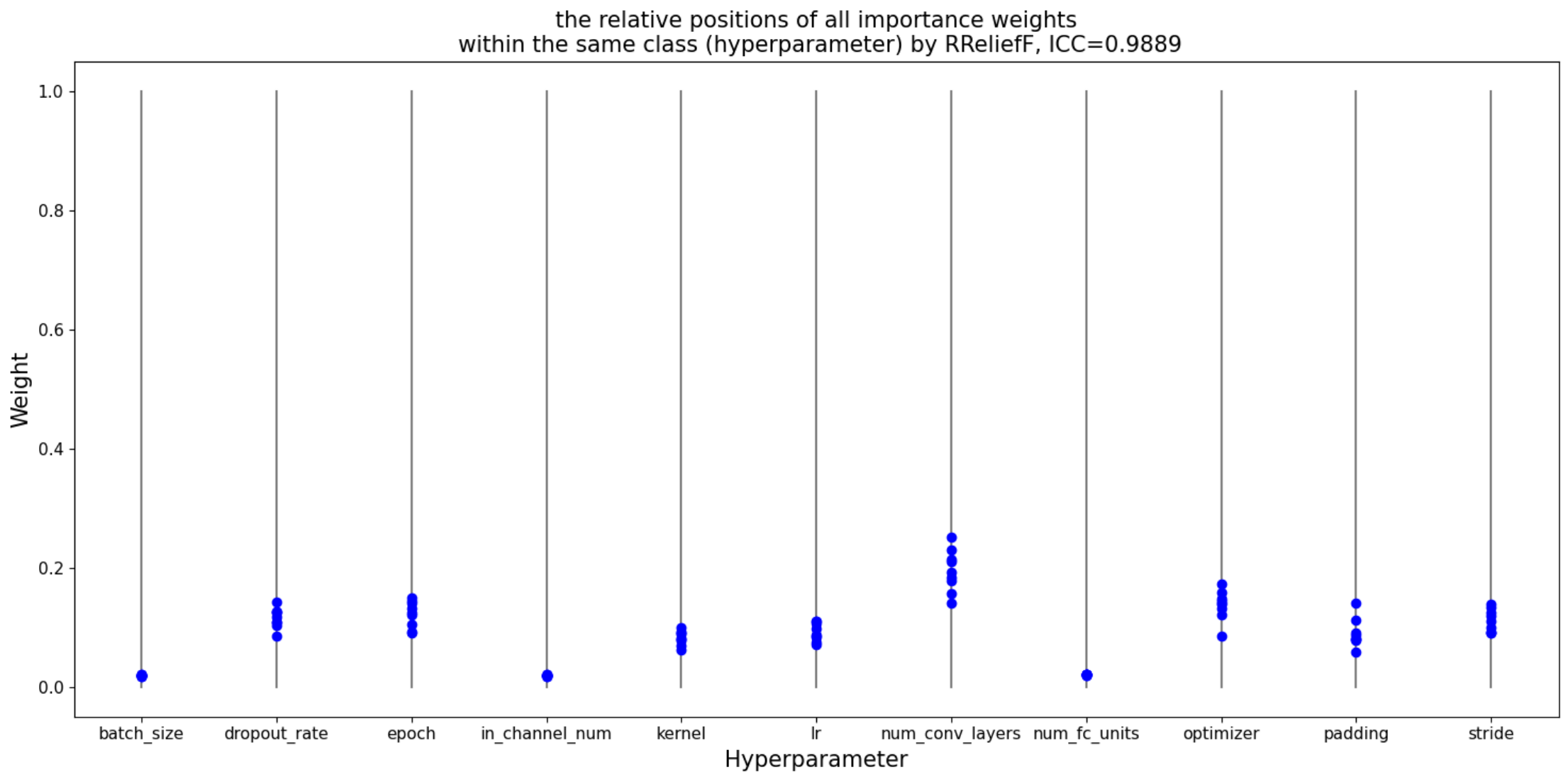}
    \label{fig:dotplot}
\end{figure}

To ensure a balanced data distribution, especially to account for the typically smaller volume of data in the medium performance intervals, a strategy of repetitive random subsampling was adopted for input into the HIA algorithm (N-RReliefF), with each performance interval limited to a maximum of 600 data points, which was considered based on the volume of data generated for each performance interval. We conducted this subsampling ten times, resulting in ten distinct subsets. Upon feeding these subsets into the HIA method, we obtained ten separate lists of hyperparameter importance weights. We then calculated the Intraclass Correlation Coefficient (ICC) to gauge the consistency of the algorithm's outputs across these different iterations. As depicted in the dot plot of Figure \ref{fig:dotplot}, there is a tight clustering of the importance weights for the same hyperparameters across the ten calculations, evidenced by an ICC of 0.9889. This high ICC value reinforces the reliability of the N-RReliefF method for assessing the hyperparameter importance of CNN.

\subsection{Importance Wights of Investigated Hyperparameters}
Finally, We executed N-RReliefF on the full dataset, setting K to 30 to ensure stable importance estimates by considering a broad set of neighbours and minimizing sensitivity to outliers. The results, shown in Table \ref{tab:results}, indicate that the number of convolutional layers, learning rate, and dropout rate are the top three most important hyperparameters in CNN models, with convolutional layers having the highest importance weight. This confirms the established view of network depth as a key factor in model performance, while learning rate and dropout rate also play significant roles in generalization and overfitting prevention. In contrast, hyperparameters such as the number of filters in fully connected layers and batch size have minimal impact on performance.

The FANOVA comparative analysis (Table \ref{tab:fanova}) corroborated these findings, revealing the same ranking of hyperparameter importance. Although the numerical weights differ, both methods highlight the critical influence of the number of convolutional layers, learning rate, and dropout rate on CNN performance.

\begin{table}[h!]  
\centering
\begin{minipage}{.4\linewidth}
\centering
\caption{Importance Weights of Investigated Hyperparameters}
\begin{adjustbox}{max width=\linewidth}
\begin{tabular}{|c|c|c|}   
\hline
{\textbf{Hyperparameter}} & {\textbf{Weights}} & {\textbf{Rank}}\\  
\hline
num\_conv\_layers & 0.385284 & 1\\ 
\hline
lr & 0.227982 & 2\\ 
\hline
dropout\_rate & 0.130576 & 3\\  
\hline
optimizer & 0.042302 & 4\\  
\hline
epoch & 0.042060 & 5\\ 
\hline
stride & 0.042030 & 6\\ 
\hline
in\_channel\_num & 0.034890 & 7\\  
\hline
padding & 0.032672 & 8\\  
\hline
kernel & 0.028568 & 9\\ 
\hline
num\_fc\_units & 0.018513 & 10\\
\hline
batch\_size & 0.015124 & 11\\ 
\hline
\end{tabular}   
\end{adjustbox}
\label{tab:results}
\end{minipage}%
\hspace{.04\linewidth}
\begin{minipage}{.4\linewidth}
\centering
\caption{Results of Comparative Analysis via FANOVA}
\begin{adjustbox}{max width=\linewidth}
\begin{tabular}{|c|c|c|}
\hline
\textbf{Hyperparameter} & \textbf{Weights} & \textbf{Rank} \\
\hline
num\_conv\_layers & 0.767432 & 1 \\
\hline
lr & 0.102461 & 2 \\
\hline
dropout\_rate & 0.081064 & 3 \\
\hline
optimizer & 0.019067 & 4 \\
\hline
epoch & 0.013576 & 5 \\
\hline
stride & 0.010516 & 6 \\
\hline
in\_channel\_num & 0.008878 & 7 \\
\hline
padding & 0.007862 & 8 \\
\hline
kernel & 0.005545 & 9 \\
\hline
num\_fc\_units & 0.005563 & 10 \\
\hline
batch\_size & 0.004528 & 11 \\
\hline
\end{tabular}
\end{adjustbox}
\label{tab:fanova}
\end{minipage}
\end{table}

\subsection{Joint Importance of Hyperparameter Pairs}
Due to the large number of hyperparameter combinations involved in joint importance, the ranking results are shown only for the top ten. Table \ref{tab:long} further proves the pivotal role that the architecture's depth plays in determining the performance of Convolutional Neural Network (CNN) models.

\begin{table}[h!]   
\caption{\textbf{Joint Importance between Every Two Hyperparameters}}
\begin{adjustbox}{scale = 0.75, center}
\begin{tabular}{|c|c|c|}   
\hline 		{\textbf{Hyperparameter}} & 
			{\textbf{Weights}}&
			{\textbf{Rank}}\\  
			\hline   (num conv layers, dropout rate) & 0.679271 & 1\\ 
			\hline  (num conv layers, optimizer) & 0.616227 & 2\\ 
			\hline  (num conv layers, epoch) & 0.564162 & 3\\  
			\hline   (num conv layers, stride)	 & 0.564025 & 4\\  
			\hline   (num conv layers, in channel num) & 0.564008 & 5\\ 
			\hline  (num conv layers, padding)	 & 0.559995 & 6\\ 
			\hline  (num conv layers, lr)	& 0.558755 & 7\\  
			\hline   (num conv layers, kernel) & 0.556466 & 8\\  
			\hline  (num conv layers, batch size) & 0.550899 & 9\\ 
			\hline   (num conv layers, num fc units) & 0.549035 & 10\\
			\hline
\end{tabular}   
\label{tab:long}
\end{adjustbox}
\end{table}

\subsection{Importance of Filter Counts Across Convolutional Layers}
There are often dependencies between hyperparameters affecting network structures. For example, when the number of convolutional layers is 3, the number of filters in the 3rd convolutional layer and the number of filters in the 2nd layer have to be set. But if the number of convolutional layers is 1, the above two hyperparameters do not exist. Thus, this section will explore the importance of ranking between the hyperparameters named ``the number of filters" of different convolutional layers. 

\begin{table}[h!]   
\centering
\begin{minipage}[t]{.48\textwidth}
\centering
\caption{If the number of convolutional layers is two}
\begin{tabular}{|c|c|c|}   
\hline {\textbf{Hyperparameter}} & 
       {\textbf{Weights}}&
       {\textbf{Rank}}\\  
\hline   num of filters in layer 1 & 0.510872 & 1\\ 
\hline   num of filters in layer 2 & 0.489128 & 2\\ 
\hline
\end{tabular}   
\label{tab:layer2}
\end{minipage}
\hfill
\begin{minipage}[t]{.48\textwidth}
\centering
\caption{If the number of convolutional layers is three}
\begin{tabular}{|c|c|c|}   
\hline {\textbf{Hyperparameter}} & 
       {\textbf{Weights}}&
       {\textbf{Rank}}\\  
\hline   num of filters in layer 1 & 0.627518 & 1\\ 
\hline   num of filters in layer 2 & 0.366161 & 2\\ 
\hline   num of filters in layer 3 & 0.006321 & 3\\ 
\hline
\end{tabular}   
\label{tab:layer3}
\end{minipage}
\end{table}

In the scenario where the CNN comprises two convolutional layers (Table \ref{tab:layer2}), the importance weights allocated to the number of filters in the first layer (0.510872) slightly exceed those in the second layer (0.489128). This suggests a marginal yet notable preference for the configuration of the initial layer over the subsequent one in terms of its influence on the model's performance.

This trend becomes more pronounced when the network depth is increased to three convolutional layers (Table \ref{tab:layer3}). Here, the importance weight of the number of filters in the first layer (0.627518) significantly surpasses those in the second (0.366161) and third layers (0.006321), underscoring a clear pattern where filters in layers closer to the input exhibit a greater impact on the model's effectiveness.

\section{Conclusions and Future Work}
\label{sec: conclusions}
In this study, we investigated whether N-RReliefF, as an HIA method, can be effectively applied in the domain of deep learning. Our analysis spanned the training of over 10,000 CNN models across a diverse spectrum of 10 image classification datasets, generating an extensive collection of hyperparameter configurations and their impact on model performance. To ensure the reliability of N-RReliefF, we computed the Intraclass Correlation Coefficient (ICC) across 10 distinct subsets from our dataset. We also undertook a comparative analysis using FANOVA. Although there were numerical variations in the importance weights, the ranking order of the hyperparameters remained consistent, which confirmed the robustness of our findings.

Our analysis revealed that the number of convolutional layers, learning rate, and dropout rate emerge as the most influential hyperparameters, in line with the established best practices observed by machine learning practitioners. This not only validates the commonly used rules of thumb in the field but also provides a quantitative basis for them, enhancing their reliability and applicability in optimizing CNN models. Additionally, our findings regarding the relative importance of filters in convolutional layers illustrate a clear trend: hyperparameters associated with layers closer to the input are more influential, supporting the principle that early layers in a network play a more critical role in performance outcomes.

While this study offers valuable insights into hyperparameter importance in CNNs, there are areas for further improvement. Future work will focus on applying HIA to renowned CNN architectures such as LeNet, AlexNet, GoogleNet, and ResNet, broadening the scope of the investigation to encompass a wider range of deep learning models and providing a more comprehensive understanding of HIA's applicability and effectiveness.

\section*{Acknowledgements}
RW gratefully acknowledges financial support from China Scholarship Council. And MG thanks EPSRC for the support on grant EP/X001091/1.

\bibliographystyle{splncs04}
\bibliography{bibliography.bib}
\end{document}